\documentclass[sigconf]{acmart}
\AtBeginDocument{%
  }

\setcopyright{acmlicensed}

\copyrightyear{2025}
\acmYear{2025}
\setcopyright{cc}
\setcctype{by}
\acmConference[MM '25]{Proceedings of the 33rd ACM International Conference on Multimedia}{October 27--31, 2025}{Dublin, Ireland}
\acmBooktitle{Proceedings of the 33rd ACM International Conference on Multimedia (MM '25), October 27--31, 2025, Dublin, Ireland}\acmDOI{10.1145/3746027.3762065}
\acmISBN{979-8-4007-2035-2/2025/10}

\usepackage{makecell}
\usepackage{multirow}
\usepackage{balance}

\begin{document}

\title{MEGC2025: Micro-Expression Grand Challenge on Spot Then Recognize and Visual Question Answering}

\author{Xinqi Fan}
\authornote{Authors contributed equally to this research.}
\orcid{0000-0002-8025-016X}
\affiliation{%
  \institution{Department of Computing and Mathematics, Manchester Metropolitan University}
  \city{Manchester}
  \country{UK}
  }
  \email{X.Fan@mmu.ac.uk}
  
\author{Jingting Li}
\authornotemark[1]
\orcid{0000-0001-8742-8488}
\affiliation{%
\institution{State Key Laboratory of Cognitive Science and Mental Health, Institute of Psychology, Chinese Academy of Sciences}
\city{}
\country{}}
\affiliation{%
\institution{Department of Psychology, University of the Chinese Academy of Sciences}
\city{Beijing}
\country{China}}
\email{lijt@psych.ac.cn}
%

\author{John See}
\orcid{0000-0003-3005-4109}
\affiliation{%
 \institution{School of Mathematical and Computer Sciences,\\ Heriot-Watt University Malaysia}
 \city{Putrajaya}
 \country{Malaysia}}
 \email{J.See@hw.ac.uk}

\author{Moi Hoon Yap}
\orcid{0000-0001-7681-4287}
\affiliation{%
  \institution{Department of Computing and Mathematics, Manchester Metropolitan University}
  \city{Manchester}
  \country{UK}}
\email{m.yap@mmu.ac.uk}

\author{Wen-Huang Cheng}
\orcid{0000-0002-4662-7875}
\affiliation{%
  \institution{National Taiwan University}
  \city{Taipei}
  \country{Taiwan}
}
\email{wenhuang@csie.ntu.edu.tw}

\author{Xiaobai Li}
\orcid{0000-0003-4519-7823}
\affiliation{%
\institution{Zhejiang University}
\city{Hangzhou}
\country{China}}
\affiliation{%
\institution{University of Oulu}
\city{Oulu}
\country{Finland}}
\email{xiaobai.li@zju.edu.cn}

\author{Xiaopeng Hong}
\orcid{0000-0002-0611-0636}
\affiliation{%
  \institution{Harbin Institute of Technology}
  \city{Harbin}
  \country{China}}
 \email{hongxiaopeng@ieee.org}

\author{Su-Jing Wang}
\orcid{0000-0002-8774-6328}
\affiliation{%
\institution{State Key Laboratory of Cognitive Science and Mental Health, Institute of Psychology, Chinese Academy of Sciences}
\city{}
\country{}}
\affiliation{%
\institution{Department of Psychology, University of the Chinese Academy of Sciences}
\city{Beijing}
\country{China}}
\email{wangsujing@psych.ac.cn}

\author{Adrian K. Davison}
\authornotemark[1]
\orcid{0000-0002-6496-0209}
\affiliation{%
  \institution{Department of Computing and Mathematics, Manchester Metropolitan University}
  \city{Manchester}
  \country{UK}
  }
  \email{A.Davison@mmu.ac.uk}

\renewcommand{\shortauthors}{Xinqi Fan et al.}

\begin{abstract}
  Facial micro-expressions (MEs) are involuntary movements of the face that occur spontaneously when a person experiences an emotion but attempts to suppress or repress the facial expression, typically found in a high-stakes environment. In recent years, substantial advancements have been made in the areas of ME recognition, spotting, and generation. However, conventional approaches that treat spotting and recognition as separate tasks are suboptimal, particularly for analyzing long-duration videos in realistic settings. Concurrently, the emergence of multimodal large language models (MLLMs) and large vision-language models (LVLMs) offers promising new avenues for enhancing ME analysis through their powerful multimodal reasoning capabilities. The ME grand challenge (MEGC) 2025 introduces two tasks that reflect these evolving research directions: (1) ME spot-then-recognize (ME-STR), which integrates ME spotting and subsequent recognition in a unified sequential pipeline; and (2) ME visual question answering (ME-VQA), which explores ME understanding through visual question answering, leveraging MLLMs or LVLMs to address diverse question types related to MEs. All participating algorithms are required to run on this test set and submit their results on a leaderboard. More details are available at \url{https://megc2025.github.io}.
\end{abstract}





\maketitle

\section{Introduction}
\begin{figure*}[!t]
    \centering
    \includegraphics[width=0.8\textwidth]{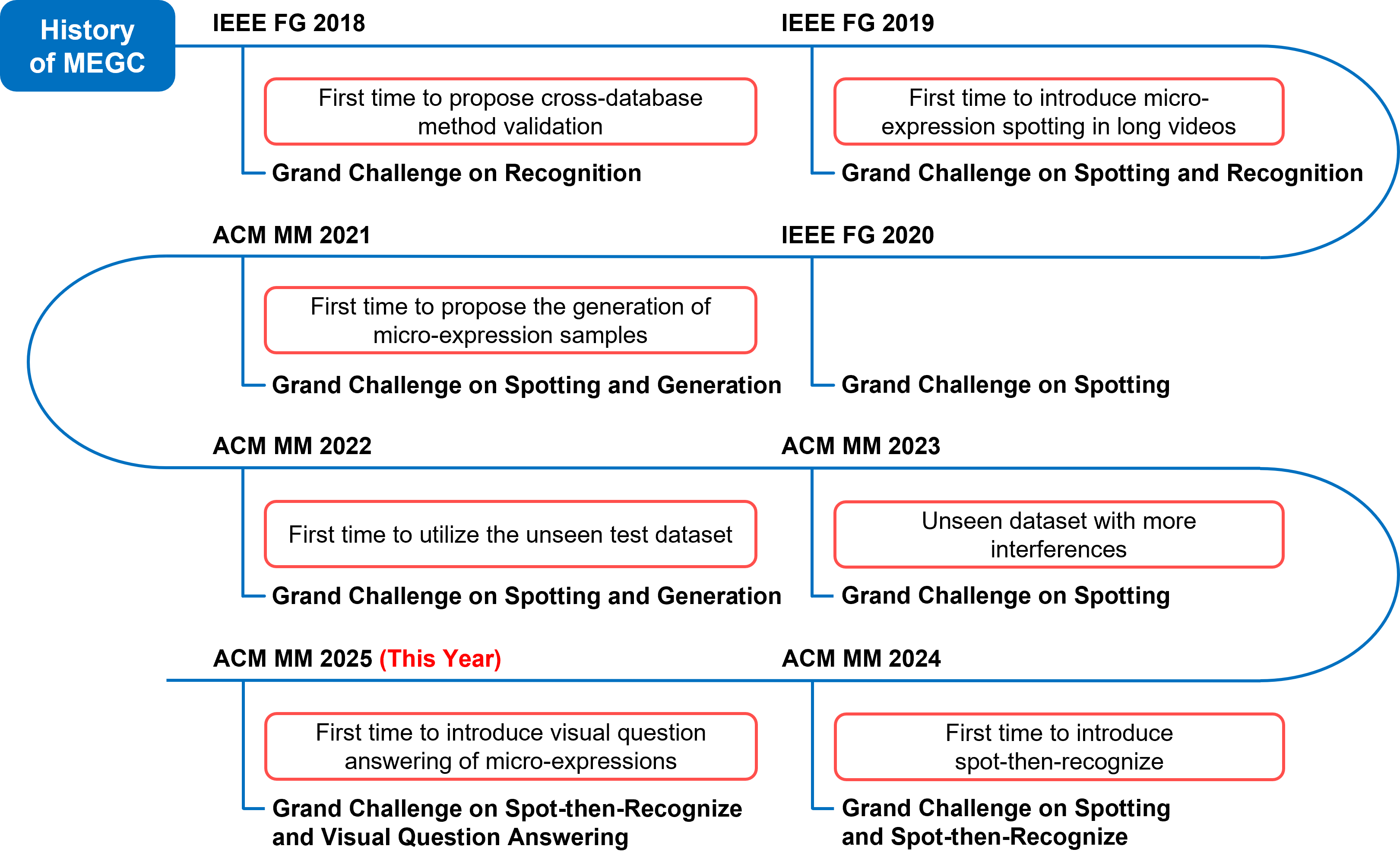}
    \caption{An overview of Micro-Expression Grand Challenges (MEGCs).}
    \label{fig:megc_history}
\end{figure*}
When a person attempts to suppress a facial expression, typically in a high-stakes scenario, there is a possibility of an involuntary movement occurring on the face, namely a facial micro-expression (ME)~\cite{ben2021video}. As such, the duration of an ME is very short, generally being no more than 500 milliseconds (ms), and is the telltale sign that distinguishes them from a normal facial expression~\cite{li2022cas}. Computational analysis and automation of tasks on MEs is an emerging area in multimedia research. However, only recently, the availability of a few spontaneously induced facial ME datasets has provided the impetus to advance further from the computational aspect.

Over the years, we have organised a series of ME grand challenges (MEGCs) to explore emerging directions and advance the state-of-the-art in ME research. To date, seven MEGCs have been held: FG’18~\cite{yap2018facial}, focusing on ME recognition; FG’19~\cite{see2019megc}, addressing both ME spotting and recognition; FG’20~\cite{li2020megc}, centered on ME spotting; MM’21~\cite{li2021fme}, introducing ME generation and continuing with spotting; MM’22~\cite{li2022megc2022}, continuing with spotting and generation; MM’23~\cite{davison2023megc2023}, with a focus on spotting; and MM’24~\cite{see2024megc2024}, which introduced the spot-then-recognize paradigm and cross-cultural spotting. An overview of all the MEGCs can be found in Fig.~\ref{fig:megc_history}.

This year marks the 8th MEGC, held in conjunction with ACM Multimedia 2025. Two tasks are featured: (1) ME spot-then-recognize (ME-STR), and (2) ME visual question answering (ME-VQA).

\textbf{Task 1: ME-STR.} Since the surge of interest in ME research over the past decade, the community has primarily treated spotting and recognition as separate tasks. However, such separation is often impractical in real-world settings. ME recognition alone assumes that the ME sequence has already been spotted, which is rarely the case in the real world. Conversely, ME spotting without recognition fails to provide a meaningful interpretation of the emotional content. The spot-then-recognize (STR) paradigm offers a more realistic setting by first identifying the occurrence of an ME (spotting), and then classifying it into an ME category (recognition). In this task, only correctly spotted samples are passed on for emotion classification. This task is a continuation from MEGC 2024~\cite{see2024megc2024}.

\textbf{Task 2: ME-VQA.} The rise of multimodal large language models (MLLMs) and large vision-language models (LVLMs) has opened new opportunities for ME analysis through their advanced multimodal reasoning capabilities. In this task, traditional ME annotations, such as emotion categories and action units, are reformulated into question-answer (QA) pairs. Given an image or video sequence along with a natural language prompt, the model is expected to generate answers that describe the observed MEs, their attributes, and related information. This task introduces a novel multimodal direction for ME understanding, promoting interpretability, flexibility, and human-aligned interaction through natural language. ME-VQA is a new task introduced in MEGC 2025.

\begin{table*}[t!]
    \centering
    \caption{Baseline results of the ME-STR task. The test set results of the ME-STR task using MEAN (the baseline method) trained on CAS(ME)$^2$. STRS refers to the spot-then-recognize score.}
    \begin{tabular}{cccccccccccccccc}
        \hline
        & \multicolumn{6}{c}{Spotting} 
        & \multicolumn{8}{c}{Recognition} 
        & \multirow{3}{*}{STRS} \\
        \cmidrule(lr){2-7} \cmidrule(lr){8-15}     
         & TP & FP & FN & Precision & Recall & F1 & TP & FP & FN & Precision & Recall & F1 & UF1 & UAR  \\
        \hline
        SAMM & 6 & 217 & 202 & 0.0269 & 0.0288 & 0.0278 & 1 & 1 & 1 & 0.1667 & 0.3333 & 0.2222 & 0.2222 & 0.1667 & 0.0062 \\
        CAS(ME)$^3$ & 3 & 115 & 34 & 0.0254 & 0.0811 & 0.0387 & 1 & 1 & 1 & 0.3333 & 1.667 & 0.2222 & 0.2222 & 0.3333 & 0.0086\\
        \hline
    \end{tabular}
    \label{tab:str_results}
\end{table*}

\begin{table*}[t!]
    \centering
    \caption{Challenge evaluation of the ME-STR task. The test set results of the ME-STR task from the top-performing teams. STRS refers to the spot-then-recognize score.}
    \begin{tabular}{cccccccc}
    \hline
    \multirow{2}{*}{Participant} & \multicolumn{2}{c}{SAMM (F1)} & \multicolumn{2}{c}{CAS(ME)$^3$ (F1)} & \multicolumn{3}{c}{STRS} \\
    \cmidrule(lr){2-3} \cmidrule(lr){4-5} \cmidrule(lr){6-8} &
    Spotting & Recognition &  Spotting & Recognition & SAMM & CAS(ME)$^3$ & Overall \\ 
    \hline
    \makecell{Guo~et~al.~\cite{guo2025boosting} \\ (guoguo01)}   & 0.086 & \textbf{0.667} & \textbf{0.099} & \textbf{0.645} & \textbf{0.057} & \textbf{0.064} & \textbf{0.09} \\ 
    \hline
    ustc-iat & \textbf{0.118} & 0.471 & 0.067 & \textbf{0.645} & 0.055 & 0.043 & 0.06 \\ 
    \hline
    gormanv & 0.067 & 0.622 & 0.061 & 0.278 & 0.041 & 0.017 & 0.047 \\ 
    \hline
    \end{tabular}
    \label{tab:str_winners}
\end{table*}

\section{Task 1: Spot-Then-Recognize}
\subsection{Task Formulation}
Much work on ME analysis has focused on two separate tasks: spotting and recognizing. Only recognizing an ME class can be unrealistic in a real-world setting, as it assumes that the ME sequence has already been identified. On the other hand, the spotting task is unrealistic in its applicability since it cannot interpret the actual emotional state of the person observed. A more realistic setting, also known as spot-then-recognize (STR), performs spotting followed by recognition in a sequential manner. Only samples that have been correctly spotted in the spotting step will be passed on to the recognition step to be classified into their emotion class.

\subsection{Dataset}
We do not have a restriction on the training set, while we recommend participants to use SAMM-LV~\cite{yap2020samm}, CAS(ME)$^3$~\cite{li2022cas}, 4DME~\cite{li20224DME}, CAS(ME)$^2$~\cite{qu2017cas}, or SMIC-E-long~\cite{tran2021micro} datasets for training.

For the ME-STR test set, it contains 30 long videos, including 10 long videos from the SAMM Challenge dataset \cite{yap20223d, davison2018samm} and 20 clips cropped from different videos in CAS(ME)$^3$ (unreleased before). The frame rate for the SAMM Challenge dataset is 200fps, and the frame rate for CAS(ME)$^3$ is 30 fps. Participants should test on this unseen dataset.

\subsection{Evaluation Metrics}
Several performance metrics are used in this work to ensure fair comparisons and to provide a comprehensive evaluation of the ME spotting and recognition tasks. For ME spotting, true positive (TP), false positive (FP), false negative (FN), precision, recall, and F1 are used. For ME recognition, TP, FP, FN, precision, recall, F1, unweighted F1 (UF1), and unweighted (UAR) are used. Finally, to comprehensively evaluate the performance of the complete ME analysis system, the spot-then-recognize score (STRS) is employed as the product of the F1-scores of spotting ($\mathrm{F1\text{-}score}_s$) and analysis ($\mathrm{F1\text{-}score}_a$):
\begin{equation}
\mathrm{STRS} = \mathrm{F1\text{-}score}_s \times \mathrm{F1\text{-}score}_a.
\end{equation}
The details of these metrics can be found in \cite{liong2023spot}.

For our baseline method, we include detailed evaluation metrics. To simplify the challenge evaluation, the methods from the challenge participants will be mainly evaluated based on $\mathrm{F1\text{-}scores}$ and STRS.

\subsection{Base                                                                                                                                      line Method}
We used the micro-expression analysis network (MEAN)~\cite{liong2023spot} as our baseline method. MEAN is a unified neural network architecture designed to perform both ME spotting and recognition in an end-to-end manner. It is structured with shared layers followed by two distinct branches: one for spotting and one for recognition. The spotting branch functions as a regression branch, outputting a confidence score for each frame to indicate the likelihood of an ME interval. Once the spotting branch identifies candidate intervals of MEs, the recognition branch predicts the emotion class for each interval.

\subsection{Baseline Results}
Table~\ref{tab:str_results} presents the quantitative results of the STR task using the MEAN network training on CAS(ME)$^2$ and testing on the MEGC 2025 ME-STR test set. The spotting results indicate the substantial challenge of localising MEs in long video sequences, as evidenced by the high number of FPs and FNs, and the low numbers of precision, recall, and F1 in both datasets. The recognition results, evaluated on the intervals detected by the spotting module, demonstrate slightly higher performance on certain metrics, but are ultimately constrained by the quality of the detected intervals. The overall STRS score, defined as the product of the spotting and recognition F1-scores, remains low for both datasets (0.0062 for SAMM and 0.0086 for CAS(ME)\textsuperscript{3}), primarily due to the limited recall and precision achieved in the spotting stage. This outcome highlights that the bottleneck in ME analysis systems lies predominantly in the spotting task; improvements in the temporal localisation of ME are likely to yield substantial gains in end-to-end analysis performance.

\subsection{Challenge Evaluation}
This year, we had 47 registered participants, with 8 participants submitting to the leaderboard for the ME-STR task. The test set results of the ME-STR task from the top-3 teams are shown in Table~\ref{tab:str_winners}.

Guo~et~al.~(guoguo01)~\cite{guo2025boosting} introduces a prior-guided video-level regression framework for the ME-STR task, which achieved 1st place on the ME-STR task. It aims to address issues with manually selecting predefined windows for classification and analysis of MEs. The framework integrates scalable interval selection and synergistic optimisation to improve both spotting and recognition of MEs in long video sequences. It also introduces an optimisation framework that works to create synergy between spotting and recognition. This method achieved an overall STRS score of 0.09, and also achieved the best results in almost every other metric for the individual datasets.

The participant ustc-iat introduces frequency-gated spatiotemporal learning (FGSL), a novel framework for ME-STR, achieving the second place in terms of overall STRS. FGSL contains 3 core contributions: a spatiotemporal information gating adapter (IGA) to enhance dependency modelling and attempt to prevent overfitting, a pyramid-based task decoupling to separate localisation and classification, and finally a frequency-based information filtering module (FIFM) for selective feature retention in different scales. Although this method achieved 2nd place overall, it achieved the highest F1 score for spotting on the SAMM dataset. In addition, the method tied top for the analysis F1-score on CAS(ME)$^3$.

The participant gormanv introduces a unified framework for ME analysis using a spatiotemporal multimodal large language model (STM-LLM). It achieved 3rd place in the ME-STR task. It aims to overcome the traditional separation of spotting and recognition tasks by leveraging MLLMs to jointly analyse emotional changes between two frames in video sequences. The method achieved an overall STRS score of 0.047. It also showed promise with the second-highest F1-score on the SAMM analysis.

\section{Task 2: Visual Question Answering}
\subsection{Task Formulation}
In the ME-VQA setting, the input to the model consists of (i) a sequence of video frames capturing subtle facial movements associated with MEs, and (ii) a corresponding natural language question related to the emotions or attributes of the sequence. The model is required to generate an answer in natural language, drawing on its understanding of both the visual cues and the question of MEs. These questions can cover a wide range of attributes, from binary classification such as "Is the action unit lip corner depressor shown on the face?" to multiclass classification like "What is the expression class?", and to more complex inquiries like "What are the action units present, and based on them, what is the expression class?".

\begin{table*}[t]
    \footnotesize
    \centering
    \caption{Baseline results of the ME-VQA task. The test set results of the ME-VQA task using Qwen2.5VL-3B (the baseline method) trained on the curated ME-VQA dataset. The coarse emotion class refers to positive, negative, and surprise. The fine-grained emotion class refers to happiness, surprise, fear, disgust, anger, and sadness. OAO refers to the input with onset, apex, and offset frames. OF refers to the optical flow input computed between the onset and apex frames. ZS refers to zero-shot performance, while FT refers to fine-tuned performance.}
    \begin{tabular}{@{}ccccccccccccccccccc@{}}
        \hline
        \multirow{4}{*}{Method} 
        & \multicolumn{6}{c}{SAMM} 
        & \multicolumn{6}{c}{CAS(ME)$^3$} 
        & \multicolumn{6}{c}{Overall} \\
        \cmidrule(lr){2-7} \cmidrule(lr){8-13} \cmidrule(lr){14-19} 
        & \multicolumn{2}{c}{Coarse} 
        & \multicolumn{2}{c}{Fine-Grained} 
        &  & 
        & \multicolumn{2}{c}{Coarse} 
        & \multicolumn{2}{c}{Fine-Grained} 
        &  & 
        & \multicolumn{2}{c}{Coarse} 
        & \multicolumn{2}{c}{Fine-Grained} 
        &  & 
        \\        
         & UF1 & UAR & UF1 & UAR & BLEU & ROUGE & UF1 & UAR & UF1 & UAR & BLEU & ROUGE & UF1 & UAR & UF1 & UAR & BLEU & ROUGE  \\
        \hline
        Video (ZS) & 0.242 & 0.333 & 0 & 0 & 0 & 0.295 & 0.262 & 0.333 & 0.057 & 0.048 & 0.089 & 0.315 & 0.256 & 0.333 & 0.057 & 0.048 & 0.065 & 0.312 \\
        Video (FT) & 0.206 & 0.25 & 0.057 & 0.1 & 0.234 & 0.331 & 0.376 & 0.515 & 0.044 & 0.111 & 0.109 & 0.375 & 0.328 & 0.422 & 0.059 & 0.117 & 0.122 & 0.362 \\
        \hline
        OAO (ZS) & 0.267 & 0.333 & 0.000 & 0.000 & 0.031 & 0.332 & 0.438 & 0.444 & 0.000 & 0.000 & 0.038 & 0.322 & 0.366 & 0.400 & 0.000 & 0.000 & 0.036 & 0.328 \\
        OAO (FT) & 0.300 & 0.417 & 0.111 & 0.167 & 0.182 & 0.366 & 0.401 & 0.455 & 0.054 & 0.143 & 0.021 & 0.323 & 0.396 & 0.461 & 0.061 & 0.143 & 0.027 & 0.339 \\
        \hline
        OF (ZS) & 0.3 & 0.333 & 0.19 & 0.3 & 0 & 0.359 & 0.3 & 0.455 & 0.107 & 0.143 & 0.009 & 0.263 & 0.3 & 0.4 & 0.119 & 0.16 & 0.008 & 0.289 \\
        OF (FT) & 0.489 & 0.5 & 0.067 & 0.1 & 0 & 0.449 & 0.286 & 0.323 & 0.128 & 0.167 & 0.092 & 0.355 & 0.343 & 0.378 & 0.103 & 0.133 & 0.082 & 0.382 \\
        \hline
    \end{tabular}
    \label{tab:vqa_results}
\end{table*}

\begin{table*}[t]
    \centering
    \scriptsize
    \caption{Challenge evaluations of the ME-VQA task. The test set results of the ME-VQA task from the top-performing teams. The coarse emotion class refers to positive, negative, and surprise. The fine-grained emotion class refers to happiness, surprise, fear, disgust, anger, and sadness.}
    \begin{tabular}{@{}cc@{\quad}cc@{\quad}cccc@{\quad}cc@{\quad}cccc@{\quad}cc@{\quad}cccc@{}}
    \hline
    \multirow{3}{*}{Participant} &
    \multicolumn{6}{c}{SAMM} &
    \multicolumn{6}{c}{CAS(ME)$^3$} &
    \multicolumn{7}{c}{Overall} \\ 
    \cmidrule(lr){2-7} \cmidrule(lr){8-13} \cmidrule(lr){14-20} 
    &
    \multicolumn{2}{c}{Coarse} &
    \multicolumn{2}{c}{Fine-Grained} &
    \multicolumn{1}{c}{} &
    \multicolumn{1}{c}{} &
    \multicolumn{2}{c}{Coarse} &
    \multicolumn{2}{c}{Fine-Grained} &
    \multicolumn{1}{c}{} &
    \multicolumn{1}{c}{} &
    \multicolumn{2}{c}{Coarse} &
    \multicolumn{2}{c}{Fine-Grained} &
    \multicolumn{1}{c}{} &
    \multicolumn{1}{c}{} \\
    &
    UF1 &
    UAR &
    UF1 &
    UAR &
    BLEU &
    ROUGE &
    UF1 &
    UAR &
    UF1 &
    UAR &
    BLEU &
    ROUGE &
    UF1 &
    UAR &
    UF1 &
    UAR &
    BLEU &
    ROUGE &
    Avg \\ 
    \hline
    \makecell{Wang~et~al.~\cite{emotion2025wang} \\ (lianxin-tech)} &
    \textbf{0.778} &
    \textbf{0.833} &
    0.427 &
    0.500 &
    0.543 &
    0.625 &
    \textbf{0.717} &
    \textbf{0.717} &
    0.294 &
    \textbf{0.389} &
    \textbf{0.643} &
    \textbf{0.605} &
    \textbf{0.733} &
    \textbf{0.722} &
    \textbf{0.368} &
    \textbf{0.408} &
    0.615 &
    \textbf{0.607} &
    \textbf{0.575} \\ 
    \hline
    \makecell{Zhu~et~al.~\cite{hierMEQA2025zhu} \\ (ustc-iat)} &
    0.750 &
    0.750 &
    \textbf{0.660} &
    \textbf{0.700} &
    0.609 &
    \textbf{0.637} &
    0.561 &
    0.626 &
    0.217 &
    0.278 &
    0.589 &
    0.458 &
    0.594 &
    0.650 &
    0.316 &
    0.375 &
    0.595 &
    0.509 &
    0.506 \\ 
    \hline
    IIM, HFIPS, CAS &
    0.675 &
    0.750 &
    0.143 &
    0.143 &
    0.427 &
    0.450 &
    0.438 &
    0.444 &
    \textbf{0.338} &
    0.333 &
    0.364 &
    0.505 &
    0.560 &
    0.528 &
    0.281 &
    0.283 &
    0.396 &
    0.489 &
    0.423 \\ 
    \hline
    \end{tabular}
    \label{tab:vqa_winners}
\end{table*}

\subsection{Dataset}
We do not have a restriction on the training set, while we recommend participants to use SAMM~\cite{davison2018samm}, CASME~II~\cite{yan2014casme}, SMIC~\cite{li2013spontaneous}, CAS(ME)$^3$~\cite{li2022cas}, or 4DME~\cite{li20224DME} datasets for training.

We also convert the SAMM, CASME II, and SMIC annotations to a ME-VQA version\footnote{The ME-VQA version dataset is available at~\url{https://megc2025.github.io/challenge.html}} that is suitable to train a ME-VQA model.

The ME-VQA test set contains 24 ME clips, including 7 clips from the SAMM Challenge dataset \cite{yap20223d, davison2018samm} and 17 clips cropped from different videos in CAS(ME)$^3$ (unreleased before). The frame rate for SAMM is 200 fps, and the frame rate for CAS(ME)$^3$ is 30 fps. The participants should test on this unseen dataset.

\subsection{Evaluation Metrics}
To evaluate the performance of our ME-VQA baseline, we report metrics for both emotion classification and the overall language answer quality.

For both coarse- and fine-grained emotion classification, we use unweighted F1 Score (UF1) and Unweighted Average Recall (UAR) to ensure balanced evaluation across classes. The coarse emotion classes refer to positive, negative, and surprise. The fine-grained emotion classes refer to happiness, surprise, fear, disgust, anger, and sadness. The details of these metrics can be found in~\cite{see2019megc}.

For all VQA answers, we report bilingual evaluation understudy (BLEU)~\cite{papineni2002bleu} and recall-oriented understudy for gisting evaluation (ROUGE)-1~\cite{chin2004rouge} to assess the quality of generated text. BLEU evaluates n-gram precision between predicted and reference answers as:
\begin{equation}
    \mathrm{BLEU} = \exp \left( \min\left(1 - \frac{r}{c}, 0\right) + \sum_{n=1}^{N} w_n \log p_n \right),
\end{equation}
where $r$ is the reference length, $c$ is the candidate length, $p_n$ is the $n$-gram precision, and $w_n$ are the weights. 
The ROUGE-1 score is defined as the recall of unigram overlaps between the candidate answer $C$ and the reference answer $R$:
\begin{equation}
    \mathrm{ROUGE\text{-}1} = \frac{\sum_{w \in V} \min\big(\mathrm{N}_p(w),\ \mathrm{N}_r(w)\big)}{\sum_{w \in V} \mathrm{N}_r(w)},
\end{equation}
where $V$ is the set of unique words in the reference answer, and $\mathrm{N}_C(w)$ and $\mathrm{N}_R(w)$ are the occurrences of word $w$ in the candidate and reference answers, respectively.

The final ranking of the challenge participants will be based on the average (Avg) of all the overall metrics.

\subsection{Baseline Method}
For the ME-VQA task, we employ the Qwen2.5VL-3B model~\cite{bai2025qwen25} as our baseline method. Qwen2.5VL-3B is a recent LVLM that is based on the Qwen 2.5 architecture and incorporates a vision encoder, a language model backbone, and a cross-modal fusion module. The model is pre-trained on large-scale image-text datasets, enabling it to perform a wide range of VQA tasks with strong zero-shot and few-shot generalisation capabilities.

To establish a general baseline, we consider zero-shot (ZS) and fine-tuning (FT) of the Qwen2.5VL-3B model. For fine-tuning, we applied LoRA to the vision encoder, projection layers between vision and language, as well as the query and keys of the language model. We also investigated three input types: (1) Input with equally sampled video frames. (2) Input with onset, apex, and offset frames (OAO). (3) Input with optical flow between onset and apex (OF).

\subsection{Baseline Results}
The experimental results presented in Table~\ref{tab:vqa_results} provide baseline evaluations using Qwen2.5VL-3B on the test set of the MEGC 2025 ME-VQA task.

From the results, it is evident that the smallest 3B version of Qwen2.5-VL can perform some basic ME recognition. Fine-tuning improves performance across most metrics compared to zero-shot evaluation, but the improvement is not significant, likely due to only a small number of parameters were updated using LoRA.

For expressions, especially the coarse ones, OAO (onset-apex-offset) and OF (optical flow) inputs consistently outperform the video-input results, both in zero-shot and fine-tuned conditions. This indicates that providing explicit temporal structure or motion information enables the model to better capture the subtle dynamics inherent in ME recognition. However, video-input models have better overall answer quality in terms of BLEU and ROUGE-1.

\subsection{Challenge Evaluation}
The ME-VQA task attracted 28 participants, resulting in 10 teams being submitted to the leaderboard with 259 valid submissions. Among the 10 teams, 9 reported results that exceeded the baseline. The test set results of the ME-VQA task from the top-3 teams are shown in Table~\ref{tab:vqa_winners}.

Wang~et~al.~(lianxin-tech)~\cite{emotion2025wang} proposed Emotion-Qwen-VL and won first place in the ME-VQA task in the MEGC 2025 challenge. Emotion-Qwen-VL is a fully fine-tuned MLLM designed for the ME-VQA. Built on Qwen2.5-VL-7B, the model is optimised to understand subtle facial dynamics and generate interpretable natural language responses. The authors also introduced a rule-based AU-to-emotion mapping framework during the inference to provide a semantic bridge between low-level facial muscle activations and high-level expression labels. In addition, the authors found that training on CASME II and SAMM datasets led to better performance on the SAMM test dataset, while training on CASME II and SMIC datasets led to better performance on the CAS(ME)$^3$ test dataset.

Zhu~et~al.~(ustc-iat)~\cite{hierMEQA2025zhu} proposed HierMEQA and won second place in the ME-VQA task of the MEGC 2025 challenge. HierMEQA is a novel hierarchical framework designed for the emerging task of ME-VQA. The framework employs a multi-stage hierarchical reasoning process from coarse-grained emotion analysis, fine-grained emotion analysis, to localised action unit detection. In addition, the authors used a hybrid sampling strategy by sampling onset, apex, offset, and some random frames at the training time, while uniform sampling was used at the test time. Instead of full parameter fine-tuning, this method uses parameter-efficient fine-tuning with LoRA.

Participant IIM,~HFIPS,~CAS proposed a temporal information enhanced VLM (TIE) and won third place in the ME-VQA task in the MEGC 2025 challenge. The proposed framework is built upon Qwen2.5-VL-7B-Instruct and integrates raw video frames and optical flow maps to capture both static and dynamic facial cues. It also used structured prompts to guide the model to focus on specific regions during inference. Full-parameter supervised fine-tuning is used to adapt the model for ME-VQA. 

\section{Conclusions}
In this year’s MEGC challenge, we successfully organised two tasks: ME-STR and ME-VQA. A variety of teams participated, proposing diverse approaches that integrated MLLMs, advanced visual features, and reasoning strategies tailored for ME analysis. Participants exceeded the baseline results by significant margins in both tasks, demonstrating notable improvements in ME-STR and ME-VQA.

For the ME-STR task, the overall STRS remains relatively low, with the best-performing team achieving a score of 0.09. This highlights a substantial need for further exploration and methodological innovation in this area. In contrast, for the ME-VQA task, the best-performing team achieved an average score of 0.575, indicating strong performance. However, the ME-VQA test set for this year’s challenge was relatively small, suggesting that larger and more diverse test datasets will be necessary to better evaluate and validate the generalisation capabilities of ME-VQA methods.

In line with the requirements of ACM Multimedia, only three papers were accepted for publication, one from the STR task and two from the ME-VQA task, based on the performance of the leaderboard, the methodological quality, and the rigor of the submitted papers.

\begin{acks}
This research is supported in part by the National Natural Science Foundation of China (62476269, 62276252), the Youth Innovation Promotion Association CAS, and the Ministry of Higher Education (MOHE) Malaysia under the Fundamental Research Grant Scheme (FRGS) (FRGS/1/2022/ICT02/HWUM/02/1).
\end{acks}

\balance
\bibliographystyle{ACM-Reference-Format}
\bibliography{megc2025_final_summary_paper}










\end{document}